\documentclass[]{ceurart}

\usepackage{url}

\begin{document}

\copyrightyear{2022}
\copyrightclause{Copyright for this paper by its authors.
  Use permitted under Creative Commons License Attribution 4.0
  International (CC BY 4.0).}

\conference{SWAT4HCLS 2023: The 14th International Conference on Semantic Web Applications and Tools for Health Care and Life Sciences}

\title{Fine-Grained Named Entities for Corona News}

\author[1]{Sefika Efeoglu}[
orcid=0000-0002-9232-4840,
email=sefika.efeoglu@fu-berlin.de,
]

\address[1]{Freie Universitaet Berlin, Takustrasse 9, 14195 Berlin, Germany}

\author[1,2]{Adrian Paschke}[%
orcid=0000-0003-3156-9040,
email=paschke@inf.fu-berlin.de,
]

\address[2]{Fraunhofer FOKUS, Berlin, Germany}
\begin{abstract}
Information resources such as newspapers have produced unstructured text data in various languages related to the corona outbreak since December 2019. Analyzing these unstructured texts is time-consuming without representing them in a structured format; therefore, representing them in a structured format is crucial. An information extraction pipeline with essential tasks- named entity tagging and relation extraction- to accomplish this goal might be applied to these texts. This study proposes a data annotation pipeline to generate training data from corona news articles, including generic and domain-specific entities. Named entity recognition models are trained on this annotated corpus and then evaluated on test sentences manually annotated by domain experts evaluating the performance of a trained model. The code base and demonstration are available at ~\url{https://github.com/sefeoglu/coronanews-ner.git}.

\end{abstract}

\maketitle
\section{Introduction}
The coronavirus outbreak has started to spread worldwide from Wuhan, China, the origin of SARS-CoV-2~\footnote{This information was declared by WHO on~\url{https://www.who.int/publications/i/item/who-convened-global-study-of-origins-of-sars-cov-2-china-part}}, in late 2019. Local authorities of each country have taken crucial measures -such as tests, vaccines, and mask obligations- in indoor facilities to control the spread of the virus. The authorities give ongoing progress reports on these measures, published on their official web pages and news articles during the pandemic.

After the start of the pandemic, the Covid-19 Open Research Dataset~\footnote{ {https://www.kaggle.com/datasets/allen-institute-for-ai/CORD-19-research-challenge}} (CORD-19) challenge was declared to convert texts taken from previously published scientific papers in the corona domain into a structured format for downstream applications in March 2020~\cite{CORD_19}. Nevertheless, the applications using this corpus fail to identify recent variants of the coronavirus and generic mentions such as organizations and facilities in the news articles. This is because this corpus includes earlier published scientific papers in this domain. Thus, a new up-to-date corpus is needed to analyze all mentions in corona news articles.

This study aims to develop an annotation pipeline that generates annotated training data from newer corona news articles for named entity recognition (NER). After running the annotation pipeline, we leverage the Flair NLP framework~\cite{akbik2019flair} and SciBERT~\cite{Beltagy2019SciBERT} to train NER models on the annotated texts and then evaluate the models on test data annotated by domain experts. The corpus in this study is constructed from the corona news articles published in the German news channel ``Tagesschau''~\footnote{ ~\url{https://www.tagesschau.de/}} and are tagged with 23 entity types. The main contribution of this study is to introduce a new corpus from up-to-date corona news articles tagged with gold and silver seeds and a pre-trained NER model (OntoNotes~\footnote{The OntoNotes corpus has been constructed from various kinds of data sources to develop information extraction and retrieval application. The details are available on~\url{https://catalog.ldc.upenn.edu/LDC2013T19}}).

The rest of this paper outlines recent works about corona-related text data annotation processes and the NER approaches in the Related Works section. Afterward, the proposed annotation pipeline and the dataset are presented in the Methodology section, and then the experiments carried out are debated in the Evaluation section. Lastly, we summarize our work within this study in the Conclusion section.

\section{Related Works}
There are several previous attempts to construct a corpus in the corona domain ahead of the Covid-19 pandemic. However, previous corpora in this domain -such as CORD-19~\cite{CORD_19} and LitCovid~\cite{LitCovid_19}- generated from domain-specific journals need up-to-date information in identifying generic entities and new entities of coronavirus variants.

Wang et al. (2020) introduce a dataset from the CORD-19 corpus using gold seeds created by domain experts, UMLS KB, and NER models (spaCy and SciSpacy)~\cite{CORD_NER}. A NER model using a distant supervision approach is trained on this annotated corpus and evaluated on test sentences annotated by three domain experts. Another study using the CORD-19 corpus is ``Automated Text Evidence Mining'' supporting data-driven methods for distantly supervised NER and open information extraction~\cite{automated_evidence_2020}. Colic et al.(2020) also propose a pipeline to annotate scientific publications about the Covid-19 pandemic~\cite{colic-etal-2020-annotating}. They leverage the CRAFT corpus annotated with ten entity types from domain-specific ontologies, for example, COVoc4.

Turning to studies using corpora in low-resource languages, Truong et al. (2021) employ a Covid-19 NER model trained on Vietnamese text corpus~\cite{truong-etal-2021-covid}. The corpus with 35K train and 1K test sentences consists of fully manual annotated texts regarding the Covid-19 situation in Vietnamese.
Another study in this field aims to investigate their ability to transfer knowledge between two languages while maintaining necessary features to identify named entities in which datasets are Italian SIRM Covid-19 and English medical records~\cite{CATELLI2020106779}.

To sum up, the previous attempts to create an annotated corpus in the corona domain leverage outdated published texts. However, outdated text data is insufficient for recognizing mentions in changing corona news articles.
\section{Methodology}

\subsection{Data and Data Preprocessing}
\begin{figure}[htbp]
    \centering
    \includegraphics[height=5cm, width=11cm]{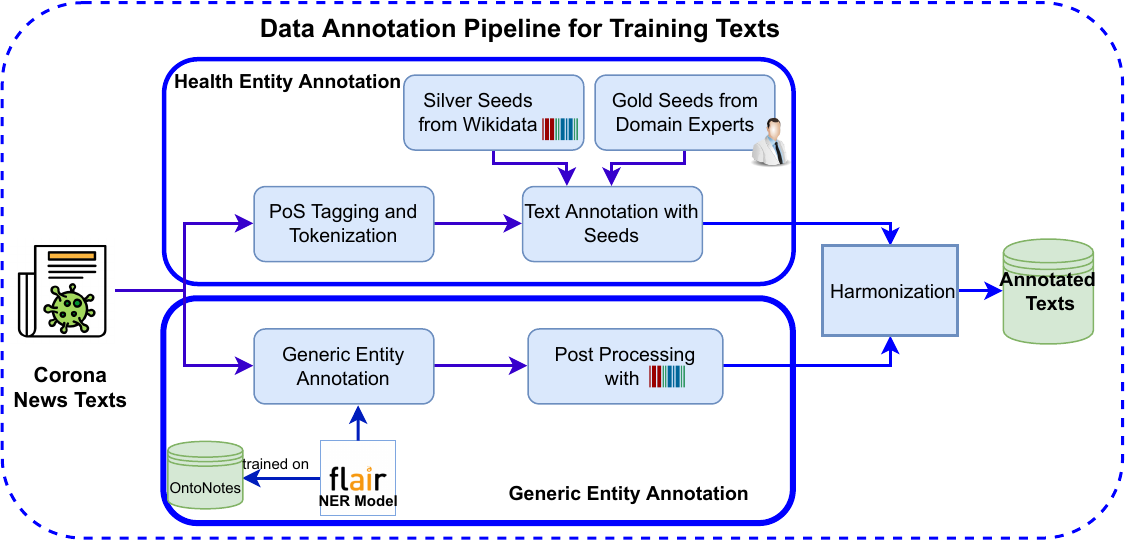}
    \caption{This figure illustrates how a corona news article is annotated with (silver and gold) seed entities and a pre-trained NER model.}
    \setlength{\belowcaptionskip}{-10pt}
    \label{fig:ner_pipeline}
\end{figure}

This study collects its data from corona-related news articles published by a German news-channel
``Tagesschau'' between December 2020 and June 2022, since the outcome of this study will be used in the relation extraction task of the information extraction pipeline later to evaluate the progress of the Covid-19 pandemic in Germany. Due to limited silver seed entities in German, we have translated sentences in these articles into English sentences with Google Translator python library~\footnote{ \url{https://pypi.org/project/deep-translator/}}. Before running the annotation pipeline (see Fig.~\ref{fig:ner_pipeline}) on the data, fundamental text cleaning approaches -e.g, removing unwanted characters like `\#' and `*' and unnecessary white spaces- have been applied. After this cleaning process, the pipeline can be applied to unstructured text data to prepare annotated text data for text analytic approaches.

\subsection{A Data Annotation Pipeline for Corona News}
This study proposes an annotation pipeline, leveraging silver and gold seeds and a pre-trained NER model (OntoNotes) to prepare annotated texts for text analytics methods like named entity tagging and relation detection between entities in a sentence. This pipeline starts with receiving unstructured text data as input. Then it simultaneously runs two different annotation processes to extract both domain-specific (Health Entity Annotation) and generic (Generic Entity Annotation) entities. Afterwards, it solves conflicts between the annotation results of both processes in the harmonization step by prioritizing the health entity annotation's results where both processes have annotated an entity in the sentence. For instance, if `Corona' in a sentence is annotated as GPE (Generic) and Coronavirus (Health), the harmonization step will resolve this conflict by assuming `Corona' as Coronavirus. The pipeline outputs annotated text data for training a NER model.

\textbf{Health Entity Annotation Process.}
The corona news articles include domain-specific entities that have not yet been categorized into generic entity types. Therefore, we create gold standard seeds with the assistance of domain experts in domain-specific categories: coronavirus, disease\_or\_syndrome, sign\_or\_symptom, and immune\_response. These gold seeds can be changed subject to the domain of the corpus before running the annotation pipeline. Furthermore, our domain experts have also provided some generic entities related to corona, for example, vaccine (product), pandemic (event), and family members (group). In addition to these gold seeds, we utilize some silver seeds in these domain-specific categories from Wikidata by running SPARQL and SKOS queries. After obtaining seeds, tokenization and part-of-speech (PoS) tagging are first applied to the corpus. Then, an exact string-matching algorithm is run with the silver and gold seeds for the tokenized sentences in the corpus. Finally, the process identifies all domain-specific named entities defined in both seed sets at the end of this process.

\textbf{Generic Entity Annotation Process.}
The corona news articles comprise generic entities like PERSON, FAC, ORG, GPE, and so on. as well, so extracting these entities is also a crucial step for analyzing the news articles. To find the generic entities, we use a NER model pre-trained on OntoNotes having 18 generic entity types~\cite{akbik2019flair}. After tagging the generic entities, these entities are sought on Wikidata with the help of SPARQL queries and refine them if found; otherwise, they will remain unmodified. Then, the process outputs annotated texts tagged with generic entities.

\section{Evaluation}
\textbf{Experimental Setup.}
We develop an annotation pipeline for corona news articles and evaluate the performances of NER models trained on these articles. The NER models implemented with the Flair framework~\cite{akbik2019flair}, utilizing two combinations of embedding types (word and contextual embeddings), and a fine-tuned SciBERT (NER) model~\cite{Beltagy2019SciBERT} use BERT transformer. The hyperparameters of the NER models implemented by Flair are a learning rate of 0.1, 10 epochs, and a batch size of 32, and those of the fine-tuned SciBERT are one epoch and a batch size of 16. The baseline model leverages only Glove word embedding~\cite{pennington2014glove}. In contrast, the advanced model has a stack of Glove word embedding and Flair contextual embedding~\cite{akbik2018coling} (in forward and backward propagation), providing the model a contextual embedding of a word in a sentence. The numbers of training, validation, and test sentences used in the development of the NER models are 89986, 4999, and 1000, respectively. Before running the pipeline, test sentences were chosen randomly from our initial corpus constructed from the news articles published on Tagesschau. Two domain experts (a medical doctor and a pharmacist) annotated 1000 sentences. Then, to ensure reliability of these test sentences, Fleiss Kappa is calculated as 0.98. These sentences consist of 3126 entities in 23 categories at the end of manual annotation.

\textbf{Results and Discussion.}
The results in Table~\ref{tab:results} show that the model using the Flair contextual embedding~\cite{akbik2018coling} outperforms the other model with only Glove word embedding~\cite{pennington2014glove} in most of the newly introduced domain-specific categories of test sentences in terms of the mean micro-F1 score of the five times trained and evaluated models. However, the model leveraging Glove embedding performs better in the `immune response', and `disease\_or\_syndrome' types, since most entities in these categories are single tokens. Moreover, we observe that these types' standard deviation (Std) is pretty high. On the other hand, the fine-tuned SciBERT model's micro F1-score is 0.7765. The entity-specific F1-scores are 0.81 (coronavirus), 0.84 (sign\_or\_symptom), 0.79 (disease\_or\_syndrome), 0.8 (immune\_response), and 0.85 (group).
\begin{table}[!htp]
    \centering
        \caption{This table shows the statistical details about mean micro-F1 scores of the NER models, which were trained and evaluated five times. Besides, the table gives the mean micro-F1 scores of new entity types on the models trained with our corona news corpus.}
            \setlength{\belowcaptionskip}{-10pt}
    \begin{tabular}
    {p{2cm}
    p{1.5cm}
    p{1.5cm}
    p{1.7cm}
    p{1.5cm}
    p{1.5cm}
    p{1.5cm}
    p{1.5cm}
    }
    \toprule 
         \rotatebox[origin=c]{30}{\textbf{Embedding}} & 
         \rotatebox[origin=c]{30}{\textbf{\shortstack{Model}}} & \rotatebox[origin=c]{30}{\textbf{\shortstack{Model\\Std}}} &
         \rotatebox[origin=c]{30}{\textbf{\shortstack{Coronavirus}}} &
         \rotatebox[origin=c]{30}{\textbf{\shortstack{Disease\\or\\ Syndrome}}}& 
         \rotatebox[origin=c]{30}{\textbf{\shortstack{Group}}}&
         \rotatebox[origin=c]{30}{\textbf{\shortstack{Immune\\Response}}}&
         \rotatebox[origin=c]{30}{\textbf{\shortstack{Sign\\or\\ Symptom}}} \\
        \midrule  
        Glove & 0.71084 & 0.003414& 0.76522&0.84152 & 0.80078 & 0.96364& 0.81922\\
        Glove+Flair &0.77162 &  0.002322 &0.78614   &0.81214  &  0.85016 &0.83264&0.86562\\
    \bottomrule
    \end{tabular}
    \label{tab:results}
 \end{table} 

\section{Conclusion}
In this study, we propose an annotation pipeline to create annotated texts from the corona news articles for NER. We also contribute with a new up-to-date annotated corpus in the corona domain to identify corona-related mentions on the corona news articles via the NER models. The experiments demonstrate that the models utilizing contextual embedding surpass the model using an only word embedding in terms of micro-F1 score. Besides, the fine-tuned SciBERT model has performed well in the domain-specific entity types. In its next version, we will integrate a spelling-checking API into this pipeline before receiving the texts, since the entities in the articles might have some spelling mistakes after translating them into English.

\begin{acknowledgments}
The research presented in this paper was supported in part by the German Federal Ministry of Education and Research (BMBF) project "PANQURA" under grant 03COV03F, in part by the European Union project "FAST-LISA" under grant 101049342.
\end{acknowledgments}
\bibliography{reference}
\end{document}